\documentclass[letterpaper, 10 pt, conference]{ieeeconf}  

\IEEEoverridecommandlockouts                              

\usepackage{graphicx}                   
\usepackage{mathptmx}                   
\usepackage{times}                      
\usepackage{amsmath}                    
\usepackage{amssymb}                    
\usepackage[boxruled,linesnumbered]{algorithm2e}
\usepackage{subfig}
\usepackage{balance}
\graphicspath{{figures/}}

\DeclareMathOperator*{\argmax}{argmax}

\newtheorem{prop}{Proposition}
\newcommand{\comment}[1]{}

\title{\LARGE \bf
Online Inverse Reinforcement Learning via Bellman Gradient Iteration
}

\author{Kun Li$^{1}$, Joel W. Burdick$^{1}$
\thanks{*This work was
supported by the National Institutes of Health, NIBIB.} \thanks{$^{1}$Kun Li and Joel W. Burdick are
with Department of Mechanical and Civil Engineering, California Institute of Technology, Pasadena,
CA 91125, USA {\tt\small kunli@caltech.edu}
}%
}
\begin{document}
\maketitle
\thispagestyle{empty}
\pagestyle{empty}
\begin{abstract}
  This paper develops an online inverse reinforcement learning algorithm aimed at efficiently
  recovering a reward function from ongoing observations of an agent's actions. To reduce the
  computation time and storage space in reward estimation, this work assumes that each observed
  action implies a change of the Q-value distribution, and relates the change to the reward function
  via the gradient of Q-value with respect to reward function parameter. The gradients are computed
  with a novel Bellman Gradient Iteration method that allows the reward function to be updated
  whenever a new observation is available. The method's convergence to a local optimum is proved.
  This work tests the proposed method in two simulated environments, and evaluates the algorithm's
  performance under a linear reward function and a non-linear reward function. The results show that
  the proposed algorithm only requires a limited computation time and storage space, but achieves an
  increasing accuracy as the number of observations grows. We also present a potential application
  to robot cleaners at home.
\end{abstract}

\section{Introduction}
\label{irl::intro}
Assuming that an agent's motion in an environment is described with a Markov Decision Process (MDP),
the agent may choose an optimal action in a given state based on the reward function of the MDP, as
described by reinforcement learning algorithms \cite{irl::rl}. However, in many cases, the agent's
actions are observable, while the reward function is hidden and needs to be estimated based on the
observed actions, hence the inverse reinforcement learning (IRL) problem. The IRL problem arises in
many applications.  For example, in robot imitation learning, inverse reinforcement learning
algorithms may learn a reward function that explains the operator's demonstrations, and use the
reward function to estimate an optimal control policy for the robot. Another application is human
motion analysis, where the reward function that explains a human's motion may also indicate
potential health problems of the subject.

Existing solutions to the IRL problem mainly work in an off-line way, by collecting a set of
observations for off-line reward estimation.  For example, the methods in \cite{irl::irl1,irl::irl2,
irl::subgradient} estimate the agent's policy from a set of observations, and estimate a reward
function that leads to the policy. The method in \cite{irl::maxentropy} collects a set of
trajectories of the agent, and estimates a reward function that maximizes the likelihood of the
trajectories. This strategy is useful in applications where the learned reward function does not
need to be updated frequently.

However, in many applications, such as long-term monitoring of human motion, the observations are
available sequentially over an infinite horizon, while the reward function may be needed
continuously. In this case, an offline solution needs to store all the past observations and
estimate the reward function whenever it is required, which is computationally infeasible. Another
scenario is employing a smart service robot that customizes its service based on each user's
preference. Such preferences cannot be modeled a priori, and the preferences may change in a long
run, thus an offline solution is infeasible.

To solve the problem, this work formulates an \textit{online inverse reinforcement learning}
algorithm, where the reward function is updated whenever a new observation is available. The
proposed method uses an initial reward function to predict the agent's action distribution, and
updates the reward function by increasing the likelihood of the observed action in the predicted
distribution. This process is repeated on every new observation, thus the reward function is learned
in an online way. This method only stores the latest observation and reward function parameters, and
updates the reward parameters once for each new observation. The parameter updating is done by
associating the reward parameter and the observation in a differentiable way via Bellman Gradient
Iteration. To the best of our knowledge, no previous work solves the inverse reinforcement learning
problem in such an online way.

The paper is organized as follows. We review existing methods on inverse reinforcement learning in
Section \ref{irl::related}, and formulate an online inverse reinforcement learning algorithm in
Section \ref{irl::irl}. We adopt a Bellman Gradient Iteration method to compute the gradients of
Q-values with respect to the reward function in Section \ref{irl::gradient}. Several experiments are
shown in Section \ref{irl::experiments}, with conclusions in Section \ref{irl::conclusions}.

\section{Related Works}
\label{irl::related}
The idea of inverse optimal control is proposed by Kalman \cite{irl::kalman}, while the Inverse
Reinforcement Learning problem is first formulated in \cite{irl::irl1}, where the agent observes the
states resulting from an assumingly optimal policy, and tries to learn a reward function that makes
the policy better than all alternatives. Since the goal can be achieved by multiple reward
functions, this paper tries to find one that maximizes the difference between the observed policy
and the second best policy. This idea is extended by \cite{irl::maxmargin}, in the name of
max-margin learning for inverse optimal control. Another extension is proposed in \cite{irl::irl2},
where the goal is not necessarily to recover the actual reward function, but to find a reward
function that leads to a policy equivalent to the observed one, measured by the total reward
collected by following that policy. These solutions cannot solve the IRL problem in an online way,
because the policy needs to be estimated from a set of observations.

Since a motion policy may be difficult to estimate from observations, a behavior-based method is
proposed in \cite{irl::maxentropy}, which models the distribution of behaviors as a maximum-entropy
model on the amount of reward collected from each behavior. This model has many applications and
extensions. For example, Nguyen et al. \cite{irl::sequence} consider a sequence of changing reward
functions instead of a single reward function. Levine et al. \cite{irl::gaussianirl} and Finn et al.
\cite{irl::guidedirl} consider complex reward functions, instead of linear ones, and use Gaussian
process and neural networks, respectively, to model the reward function. Choi et al.
\cite{irl::pomdp} consider partially observed environments, and combines a partially observed Markov
Decision Process with reward learning. Levine et al.  \cite{irl::localirl} model the behaviors based
on the local optimality of a behavior, instead of the summation of rewards.  Wulfmeier et al.
\cite{irl::deepirl} use a multi-layer neural network to represent nonlinear reward functions. These
solutions update the reward function based on the whole trajectory of an agent, thus it cannot solve
the IRL problem in an online way.

Another method is proposed in \cite{irl::bayirl}, which models the probability of a behavior as the
product of each state-action's probability, and learns the reward function via maximum a posteriori
estimation. However, due to the complex relation between the reward function and the behavior
distribution, the author uses computationally expensive Monte-Carlo methods to sample the
distribution. This work is extended by \cite{irl::subgradient}, which uses sub-gradient methods to
reduce the computations.  Another extensions is shown in \cite{irl::bayioc}, which tries to find a
reward function that matches the observed behavior. For motions involving multiple tasks and varying
reward functions, methods are developed in \cite{irl::multirl1} and \cite{irl::multirl2}, which try
to learn multiple reward functions. These solutions also depend on the policy of the agent, thus an
online extension is difficult to formulate.

The method proposed in this paper models the distribution of the state-action pairs based on
Q-values, and updates the reward for each observed action with gradient methods like
\cite{irl::subgradient}, but the proposed approach adopts two approximation methods that improve the
flexibility of action modeling, and develops a Bellman Gradient Iteration algorithm that computes
the gradient of the optimal value function and the optimal Q-function with respect to the reward
function accurately and efficiently. Besides, we show how the approximate affects the learned reward
function parameters, and give the conditions for the approximation to be more accurate. This work is
an extension of our previous work \cite{irl::BGI}.

\section{Online Inverse Reinforcement Learning}
\label{irl::irl}
\subsection{Markov Decision Process}
We assume that an agent's motion in an environment can be described with a Markov Decision Process,
defined by the following variables:
\begin{itemize}
  \item $S=\{s\}$, a set of states
  \item $A=\{a\}$, a set of actions
  \item $P_{ss'}^a$, a state transition function that defines the probability that state $s$ becomes
    $s'$ after action $a$.
  \item $R=\{r(s)\}$, a reward function that defines the immediate reward of state $s$.
  \item $\gamma$, a discount factor that ensures the convergence of the MDP over an infinite
    horizon.
\end{itemize}

Given the observed actions of the agent, inverse reinforcement learning algorithms aim to recover
the reward function that explains the actions. In long-term applications, the agent's actions are
often observed sequentially, and the reward function needs to be estimated in an online way. This
work formulates the following online inverse reinforcement learning algorithm:
\begin{itemize}
  \item given reward function $r_t$ at moment $t$, observe current state $s_t$ and predict a
    distribution of actions $P(a_t|r_t;s_t)$.
  \item observe the true action $\hat{a}_t$ taken by the agent.
  \item update the reward function $r^t$ to $r^{t+1}$ to increase the likelihood
    $P(\hat{a}_t|r_t;s_t)$ of action $\hat{a}_t$.
\end{itemize}

To model the distribution $P(a_t|r_t;s_t)$, it is necessary to relate a state-action pair $(s,a)$
with the reward function $r$. This problem can be handled in reinforcement learning algorithms by
introducing the value function $V(s)$ and the Q-function $Q(s,a)$, described by the Bellman Equation
\cite{irl::rl}:
\begin{align}
  &V^\pi(s)=\sum_{s'|s,\pi(s)}P_{ss'}^{\pi(s)}[r(s')+\gamma*V^\pi(s')],\\
  &Q^\pi(s,a)=\sum_{s'|s,a}P_{ss'}^a[r(s')+\gamma*V^\pi(s')]
\end{align}
where $V^\pi$ and $Q^\pi$ define the value function and the Q-function under a policy
$\pi:S\rightarrow A$, defined as a mapping from the set of states $S$ to the set of actions $A$,
which describes the probabilities of actions to take in a state.

For an optimal policy $\pi^*$, the value function and the Q-function should be optimized on every
state. This is described by the Bellman Optimality Equation
\cite{irl::rl}:
\begin{align}
  &V^*(s)=\max_{a\in A}\sum_{s'|s,a}P_{ss'}^a[r(s')+\gamma*V^*(s')],\\
  &Q^*(s,a)=\sum_{s'|s,a}P_{ss'}^a[r(s')+\gamma*\max_{a'\in A}Q^*(s',a')].
  \label{equation:bellmanequation}
\end{align}

\subsection{Online Reward Learning}
We assume that reward function can be expressed as a function of a finite dimensional parameter
vector $\theta_t$ to parameterize the reward function $r(s,\theta_t)$ at time $t$, and the reward
function can be a linear or non-linear function.  This work models the predicted action distribution
$P(a_t|\theta_t;s_t)$ with the method in \cite{irl::bayirl}, based on the optimal Q-value
$Q^*(s_t,a_t,\theta_t)$ under parameter $\theta_t$:
\begin{equation}
  P(a_t|\theta_t;s_t)=\frac{\exp{b*Q^*(s_t,a_t,\theta_t)}}{\sum_{\tilde{a}\in
  A}\exp{b*Q^*(s_t,\tilde{a},\theta_t)}}
  \label{equation:motionmodel}
\end{equation}
where $b$ is a parameter controlling the degree of confidence in the agent's ability to choose
actions based on Q values. For simplification of notation, in the remaining sections,
$Q(s_t,a_t,\theta_t)$ denotes the optimal Q-value of the state-action pair $(s_t,a_t)$ under reward
parameter $\theta_t$.

Under this prediction, the likelihood of the agent's action $\hat{a}_t$ is given by:
\begin{equation}
  P(\hat{a}_t|\theta_t;s_t)=\frac{\exp{b*Q(s_t,\hat{a}_t,\theta_t)}}{\sum_{\tilde{a}\in
  A}\exp{b*Q(s_t,\tilde{a},\theta_t)}}
  \label{equation:actionlikelihood}
\end{equation}
and its log-likelihood is given by:
\begin{equation}
  L(\theta_t)=b*Q(s_t,\hat{a}_t,\theta_t)-\log{\sum_{\tilde{a}\in
  A}\exp{b*Q(s_t,\tilde{a},\theta_t)}}.
  \label{equation:loglikelihood}
\end{equation}

To increase this log-likelihood, $\theta_t$ is improved by an amount proportional to the gradient of
the log-likelihood for each new observation:
\begin{equation}
  \label{equation:gradientascent}
  \theta_{t+1}=\theta_t+\alpha*\nabla L(\theta_t)
\end{equation}
where $\alpha$ is the learning rate, and the gradient of the log-likelihood is given by:
\begin{align}
  \nabla L(\theta_t)&=b*\nabla Q(s_t,\hat{a}_t,\theta_t)\nonumber\\
  &-b*\sum_{\tilde{a}\in
  A}P(\tilde{a}|\theta_t;s_t)\nabla Q(s_t,\tilde{a},\theta_t).
  \label{equation:loglikelihoodgradient}
\end{align}

To compute the gradient,  it is necessary to compute the gradient of the Q-function $\nabla
Q=\frac{\partial Q}{\partial \theta_t}=\frac{\partial Q}{\partial r_t}\cdot \frac{\partial
r_t}{\partial \theta_t}$.  The standard way to compute the optimal Q-value is with the Bellman
Equation of Optimality in Equation \eqref{equation:bellmanequation}.

However, the Q-value in Equation \eqref{equation:bellmanequation} is non-differentiable with respect
to $r_t$ or $\theta_t$ due to the max operator. Its gradient $\nabla Q(s,a,\theta_t)$ cannot be
computed in a conventional way, and the sub-gradient method in \cite{irl::subgradient} cannot
compute the gradients everywhere in the parameter space. This work introduces a method called
\textit{Bellman Gradient Iteration} to solve the problem.

\section{Bellman Gradient Iteration}
\label{irl::gradient}
To handle the non-differentiable max function in Equation \eqref{equation:bellmanequation}, this
work adopts two approximation methods. After introducing these approximations, we analyze their
approximating qualities.

\subsection{Approximation with a P-Norm Function}
The first approximation is based on a p-norm:
\begin{equation}
  \label{equation:pnorm}
  \max(a_0,\cdots,a_n)\approx (\sum_{i=0}^n a_i^k)^{\frac{1}{k}}
\end{equation}
where $k$ controls the level of approximation, and all the values $a_0,\cdots,a_n$ are assumed to be
positive. When $k=\infty$, the approximation becomes exact. In the remaining sections, this method
is called a \textit{p-norm approximation}.

Under this approximation, the Q-function in Equation \eqref{equation:bellmanequation} can be
rewritten as:
\begin{equation}
  Q_{p,k}(s,a,\theta_t,k)=\sum_{s'|s,a}P_{ss'}^a[r(s',\theta_t)+\gamma*(\sum_{a'\in
  A}Q_{p,k}(s',a',\theta_t,k))^{1/k}].
  \label{equation:qpnorm}
\end{equation}

From Equation \eqref{equation:qpnorm}, we can construct an approximately optimal value function with
p-norm approximation $V_{p,k}(s,\theta_t)$:
\begin{equation}
  V_{p,k}(s,\theta_t)=(\sum_{a\in A}Q_{p,k}(s,a,\theta_t,k))^{1/k}.
  \label{equation:vpnorm}
\end{equation}

Equations \eqref{equation:qpnorm} and \eqref{equation:vpnorm} lead to an approximate Bellman
Optimality Equation to find the approximately optimal value function and Q-function:
\begin{align}
  &Q_{p,k}(s,a,\theta_t)=\sum_{s'|s,a}P_{ss'}^a[r(s',\theta_t)+\gamma*V_{p,k}(s',\theta_t)]\label{equation:qbellpnorm},\\
  &V_{p,k}(s,\theta_t)=(\sum_{a\in A}(
  \sum_{s'|s,a}P_{ss'}^{a}[r(s',\theta_t)+\gamma*V_{p,k}(s',\theta_t)]))^k)^{1/k}\label{equation:vbellpnorm}.
\end{align}

Taking the derivative of both sides of Equation \eqref{equation:vpnorm} and Equation
\eqref{equation:qbellpnorm}, the gradients of $V_{p,k}(s,\theta_t)$ and $Q_{p,k}(s,a,\theta_t)$ with
respect to reward function parameter $\theta$ are:
\begin{align}
  &\frac{\partial V_{p,k}(s,\theta_t)}{\partial \theta_t}= \frac{1}{k}(\sum_{a\in A}Q_{p,k}(s,a,\theta_t))^{\frac{1-k}{k}}\sum_{a\in
  A}k*Q_{p,k}(s,a,\theta_t)^{k-1}*\nonumber\\
  &\qquad\qquad\qquad\frac{\partial Q_{p,k}(s,a,\theta_t)}{\partial \theta_t}\label{equation:pnormvgrad},\\
  &\frac{\partial Q_{p,k}(s,a,\theta_t)}{\partial \theta_t}=\sum_{s'|s,a}P_{ss'}^a(\frac{\partial
  r(s',\theta_t)}{\partial \theta_t}+\gamma*\frac{\partial V_{p,k}(s',\theta_t)}{\partial
  \theta_t})\label{equation:pnormqgrad}.
\end{align}

For a p-norm approximation with non-negative Q-values, the gap between the approximate value
function and the optimal value function is a function of $k$:
\[g_{p,k}(k)=(\sum_{a'\in A} Q_{p,k}(s',a',\theta_t)^k)^{\frac{1}{k}} -\max_{a'\in A}Q_{p,k}(s',a',\theta_t).\]
The gap function $g_{p,k}(k)$ describes the error of the approximation, and it has two properties.
\begin{prop}
  Assuming $Q_{p,k}(s,a,\theta_t)\geq 0, \forall s,\forall a$, the tight lower bound of
  $g_{p,k}(k)$ is zero: \[\inf_{\forall k\in \mathbb{R}} g_{p,k}(k)=0.\]
\end{prop}
\begin{prop}
  Assuming all Q-values are non-negative, $Q_{p,k}(s,a,\theta_t)\geq 0, \forall s,a$, $g_{p,k}(k)$ is
  a decreasing function with respect to increasing $k$:
  \[g'_{p,k}(k)\leq0, \forall k \in \mathbb{R}.\]
\end{prop}
The proof is given in Appendix \ref{sec:pnorm}.

\subsection{Approximation with Generalized Soft-Maximum Function}
The second approximation is based on a \textit{generalized soft-maximum function}:
\begin{equation}
  \label{equation:softmax}
  \max(a_0,\cdots,a_n)\approx\frac{\log(\sum_{i=0}^n\exp(ka_i))}{k}
\end{equation}
where $k$ controls the level of approximation. When $k=\infty$, the approximation becomes exact. The
remaining sections refer to this method as \textit{g-soft approximation}.

Under this approximation, the Q-function in Equation \eqref{equation:bellmanequation} can be
rewritten as:
\begin{equation}
  Q_{g,k}(s,a,\theta_t)=\sum_{s'|s,a}P_{ss'}^a[r(s',\theta_t)+\gamma*\frac{\log{\sum_{a'\in A}\exp (kQ_{g,k}(s',a',\theta_t))}}{k}].
  \label{equation:qgsoft}
\end{equation}

Based on Equation \eqref{equation:qgsoft}, an approximately optimal value function with g-soft
approximation takes the form:
\begin{equation}
  V_{g,k}(s,\theta_t)=\frac{\log{\sum_{a\in A}\exp (kQ_{g,k}(s,a,\theta_t))}}{k}.
  \label{equation:vgsoft}
\end{equation}

Equations \eqref{equation:qgsoft} and \eqref{equation:vgsoft} leads to an approximate
Bellman Optimality Equation to find the approximately optimal value function and Q-function:
\begin{align}
  &Q_{g,k}(s,a,\theta_t)=\sum_{s'|s,a}P_{ss'}^a[r(s',\theta_t)+\gamma*V_{g,k}(s',\theta_t)]\label{equation:qbellgsoft},\\
  &V_{g,k}(s,\theta_t)=\frac{\log{\sum_{a\in A}\exp
  (k(\sum_{s'|s,a}P_{ss'}^{a}[r(s',\theta_t)+\gamma*V_{g,k}(s',\theta_t))}}{k})\label{equation:vbellgsoft}.
\end{align}

Taking derivative of both sides of Equations \eqref{equation:vgsoft} and \eqref{equation:qbellgsoft}
yields a Bellman Gradient Equation to compute the gradients of $V_{g,k}(s,\theta_t)$ and
$Q_{g,k}(s,a,\theta_t)$ with respect to the reward function parameter $\theta$:

\begin{align}
  &\frac{\partial V_{g,k}(s,\theta_t)}{\partial \theta_t}=\sum_{a\in A}\frac{\exp (kQ_{g,k}(s,a,\theta_t))}{\sum_{a'\in A}\exp
  (kQ_{g,k}(s,a',\theta_t))} \frac{\partial Q_{g,k}(s,a,\theta_t)}{\partial
  \theta_t}\label{equation:gsoftvgrad},\\
  &\frac{\partial Q_{g,k}(s,a,\theta_t)}{\partial
  \theta_t}=\sum_{s'|s,a}P_{ss'}^a(\frac{\partial r(s',\theta_t)}{\partial
  \theta_t}+\gamma*\frac{\partial V_{g,k}(s',\theta_t)}{\partial
  \theta_t})\label{equation:gsoftqgrad}.
\end{align}

For a g-soft approximation, the gap between the approximate value function and the optimal value
function is:
\[g_{g,k}(k)=\frac{\log(\sum_{a'\in A}\exp(kQ_{g,k}(s',a',\theta_t)))}{k}-\max_{a'\in A}Q_{g,k}(s',a',\theta_t).\]
The gap has the same two properties as the \textit{p-norm approximation}.

\begin{prop}
  The tight lower bound of $g_{g,k}(k)$ is zero:
  \[\inf_{\forall k\in \mathbb{R}} g_{g,k}(k)=0.\]
\end{prop}

\begin{prop}
  $g_{g,k}(k)$ is a decreasing function with respect to increasing $k$:
  \[g'_{g,k}(k)<0, \forall k \in \mathbb{R}.\]
\end{prop}
The proof is given in Appendix \ref{sec:gsoft}.

Approximating properties of Bellman Gradient Iteration shows that the gap between the approximated
Q-value and the exact Q-value decreases with larger $k$. Thus the value of  objective function in
Equation \eqref{equation:loglikelihood} under approximation will approach the true one with larger
$k$.

Under the approximation, the objective function in Equation \eqref{equation:loglikelihood} converges
to within some range of a locally optimal value with the gradient method. Formally:
\begin{prop}
  Assuming $\theta^*$ is a local optimum of the objective function under the true Q function and
  $L(\theta^*,s,a)$ is the locally optimal value, the gradient method
  $\theta_{t+1}=\theta_t+\alpha*\nabla L(\theta_t,k,s,a)$ from a starting point in basin of a
  $\theta^*$ under the approximated gradient $\nabla L(\theta_t,k,s,a)$ will converge to
  $\theta^{k,*}$, $\forall \epsilon<<1,\exists k$ such that
  $||L(\theta^*,s,a)-L(\theta^{k,*},s,a)||<\epsilon$ and $\lim_{k\rightarrow\infty}\epsilon=0$.
\end{prop}
The proof is given in Appendix \ref{sec:convergence}.  These proofs show that the approximated
gradient converges to a point whose distance to the converged point under the true gradient is
infinitesimal with a sufficiently large approximation level $k$.

\subsection{Bellman Gradient Iteration}
Based on the Bellman Equations \eqref{equation:qbellpnorm}, \eqref{equation:vbellpnorm},
\eqref{equation:qbellgsoft}, and \eqref{equation:vbellgsoft}, we can iteratively compute the value
of each state $V(s,\theta_t)$ and the value of each state-action pair $Q(s,a,\theta_t)$ under reward
parameter $\theta_t$, as shown in Algorithm \ref{alg:value}. In the algorithm, $apprxMax$ means a
p-norm approximation of the max function for the first method, and a g-soft approximation of the max
function for the second method.

After computing the approximately optimal Q-function, with the Bellman Gradient Equation
\eqref{equation:pnormvgrad}, \eqref{equation:pnormqgrad}, \eqref{equation:gsoftvgrad}, and
\eqref{equation:gsoftqgrad}, we can iteratively compute $\frac{\partial V}{\partial \theta_t}$ and
$\frac{\partial Q(s,a,\theta_t)}{\partial \theta_t}$ with respect to the reward function parameter
$\theta_t$, as shown in Algorithm \ref{alg:gradient}.  In the algorithm, $\frac{\partial
apprxMax}{\partial Q[s,a,\theta_t]}$ corresponds to the gradient of each approximate value function
with respect to the $Q$ function, as shown in Equation \eqref{equation:pnormvgrad} and Equation
\eqref{equation:gsoftvgrad}.

In these two approximations, the value of parameter $b$ depends on an agent's ability to choose
actions based on the Q values. Without application-specific information, we choose $b=1$ as an
uninformed parameter. Given a value for parameter $b$, the motion model of the agent is defined on
the approximated Q values, where the Q-value of a state-action pair depends on both the optimal path
following the state-action pair and other paths. When the approximation level $k$ is smaller, the
Q-value of a state-action pair relies less on the optimal path, and the motion model in Equation
\eqref{equation:motionmodel} is similar to the model in \cite{irl::maxentropy}; When
$k\rightarrow\infty$, the Q-value approaches the standard Q-value, and the motion model is similar
to the model in \cite{irl::bayirl}.  By choosing different $k$ values, we can adapt the algorithm to
different types of motion models.

With empirically chosen application-dependent parameters $k$ and $b$, Algorithm \ref{alg:value} and
Algorithm \ref{alg:gradient} are used compute the gradient of each Q-value, $Q[s,a,\theta_t]$, with
respect to the reward function parameter $\theta_t$, and then the parameter is learned with the
gradient ascent method shown in Equation \eqref{equation:loglikelihood} and Equation
\eqref{equation:gradientascent}.  A multi-start strategy handles local optimum. This process is
shown in Algorithm \ref{alg:irl}.

\begin{algorithm}
  \caption{Online Inverse Reinforcement Learning}
  \label{alg:irl}
  \KwData{$S,A,P,R,\gamma$,k}
  \KwResult{Reward function}
  choose the number of random starts $n_{rs}$\;
  initialize $\theta_0=\{\theta_0^i, i=1,\cdots,n_{rs}\}$\;
  t=0\;
  \While{observation available}
  {
    observe $(s_t,a_t)$\;
    \For{$i \in range(n_{rs})$}
    {
      compute reward function based on $\theta_t^{i}$\;
      run approximate value iteration with Algorithm \ref{alg:value}\;
      run Bellman Gradient Iteration with Algorithm \ref{alg:gradient}\;
      compute gradient  $\nabla L(\theta_t^{i})$ with Equation \eqref{equation:loglikelihoodgradient}\;
      gradient ascent: $\theta_{t+1}^i=\theta_t^i+learning\_rate*\nabla L(\theta_t^i)$\;
      compute reward function based on $\theta_{t+1}^i$\;
      compute the log-likelihood based on the reward function\;
    }
    identify the reward function with the highest log-likelihood among $n_{rs}$ reward functions\;
    output the reward function\;
    t=t+1\;
  }
\end{algorithm}

\begin{algorithm}
  \caption{Approximate Value Iteration}
  \label{alg:value}
  \KwData{$S,A,P,R,\gamma$,k}
  \KwResult{optimal value $V[S,\theta_t]$, optimal action value $Q[S,A,\theta_t]$}
  assign $V[S,\theta_t]$ arbitrarily\;
  \While{$diff>threshold$}
  {
    initialize $V'[S,\theta_t]=\{0\}$\;
    \For{$s\in S$}
    {
      initialize $T[A,\theta_t]=\{0\}$\;
     \For{$a\in A$}
     {
       $T[a,\theta_t]=\sum_{s'\in S}P_{ss'}^a(R[s',\theta_t]+\gamma*V[s',\theta_t])$\;
     }
     $V'[s,\theta_t]=apprxMax(T[A],k,\theta_t)$\;
   }
    $diff=abs(V[S,\theta_t]-V'[S,\theta_t])$\;
    $V[S,\theta_t]=V'[S,\theta_t]$\;
  }
  initialize $Q[S,A,\theta_t]=\{0\}$\;
  \For{$s\in S$}
  {
    \For{$a\in A$}
    {
      $Q[s,a,\theta_t]=Q[s,a,\theta_t]+\sum_{s'\in S}P_{ss'}^a(R[s',\theta_t]+\gamma*V[s',\theta_t])$
    }
  }
\end{algorithm}

\begin{algorithm}
  \caption{Bellman Gradient Iteration}
  \label{alg:gradient}
  \KwData{$S,A,P,R,V,Q,\gamma$,k}
  \KwResult{value gradient $V_{g,k}[S,\theta_t]$, Q-value gradient $Q_{g,k}[S,A,\theta_t]$}
  assign $V_{g,k}[S,\theta_t]$ arbitrarily\;
  \While{$diff>threshold$}
  {
    initialize $V'_{g,k}[S,\theta_t]=\{0\}$\;
    \For{$s\in S$}
    {
      initialize $T_{g,k}[A,\theta_t]=\{0\}$\;
     \For{$a\in A$}
     {
       $T_{g,k}[a,\theta_t]=\frac{\partial apprxMax}{\partial Q[s,a,\theta_t]}\sum_{s'\in S}P_{ss'}^a(\frac{\partial
       R[s',\theta_t]}{\partial \theta}+\gamma*V_{g,k}[s',\theta_t])$\;
     }
     $V_{g,k}'[s,\theta_t]=\sum T_{g,k}[A,\theta_t]$\;
   }
    $diff=abs(V_{g,k}[S,\theta_t]-V'_{g,k}[S,\theta_t])$\;
    $V_{g,k}[S,\theta_t]=V'_{g,k}[S,\theta_t]$\;
  }
  initialize $Q_{g,k}[S,A,\theta_t]=\{0\}$\;
  \For{$s\in S$}
  {
    \For{$a\in A$}
    {
      $Q_{g,k}[s,a,\theta_t]=Q_{g,k}[s,a,\theta_t]+\sum_{s'\in S}P_{ss'}^a(\frac{\partial R[s',\theta_t]}{\partial \theta}+\gamma*V_{g,k}[s',\theta_t])$
    }
  }
\end{algorithm}

\section{Experiments}
\label{irl::experiments}
We test the proposed method in two benchmark environments to evaluate its accuracy, and then show a
potential application to a smart home cleaning robot.
\subsection{Benchmark Environments}
We evaluated the proposed method in two existing benchmark environments.

\begin{figure}
  \centering
  \subfloat[A testing environment: in the encircled space, only one exit exists, but the mobile
  robot can only observe the space within the dashed lines, and it has to observe the motions of
  cars, shown as black dots in the figure, to estimate the location of the
  exit.]{\includegraphics[width=0.18\textwidth]{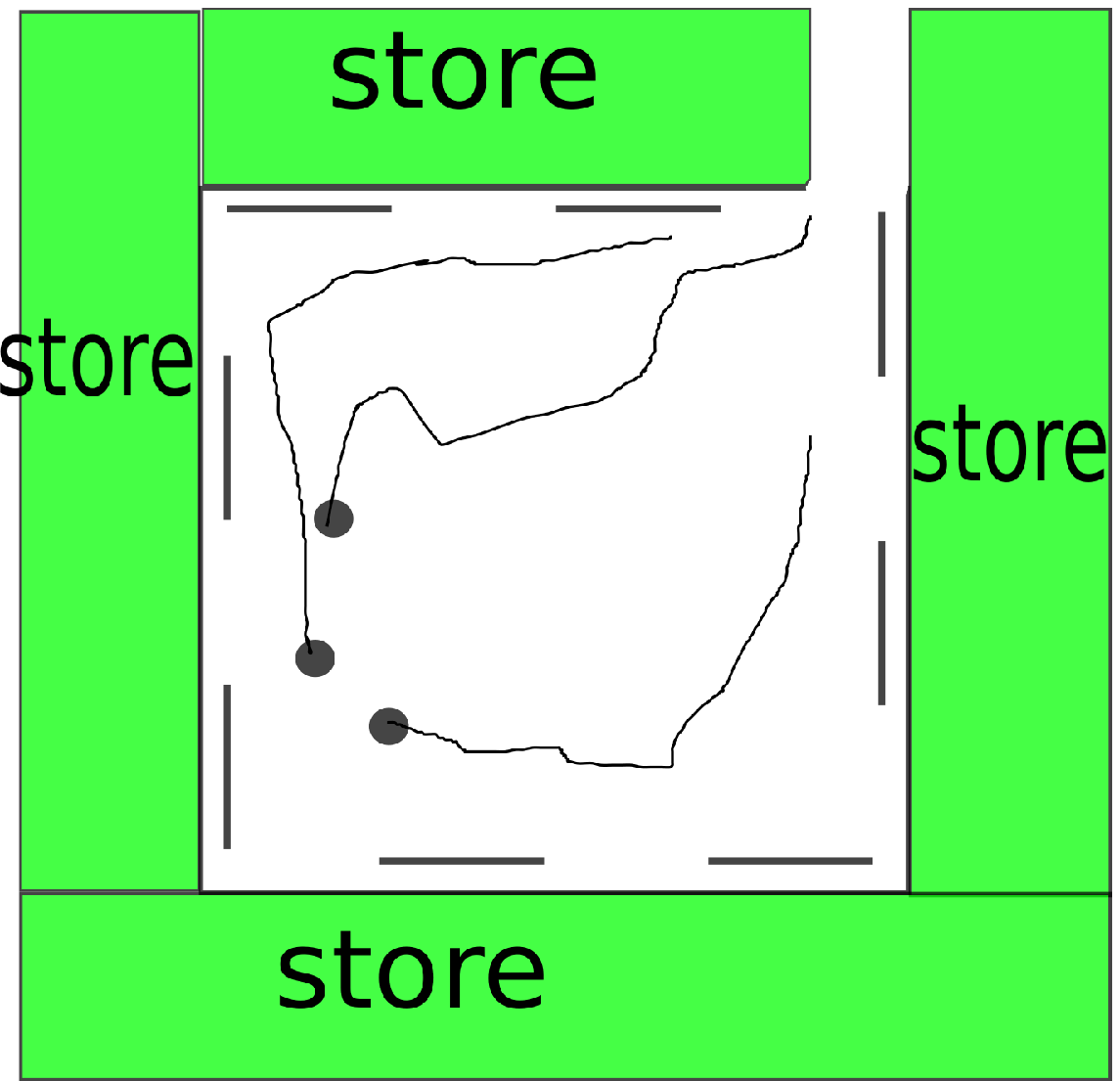}
  \label{fig:environment}}
  ~
  \subfloat[Example trajectories in Gridworld MDP: each agent starts from a random position, and
  follows an optimal policy to approach the exit. The black dots represent the initial
  positions of the agents.]{\includegraphics[width=0.22\textwidth]{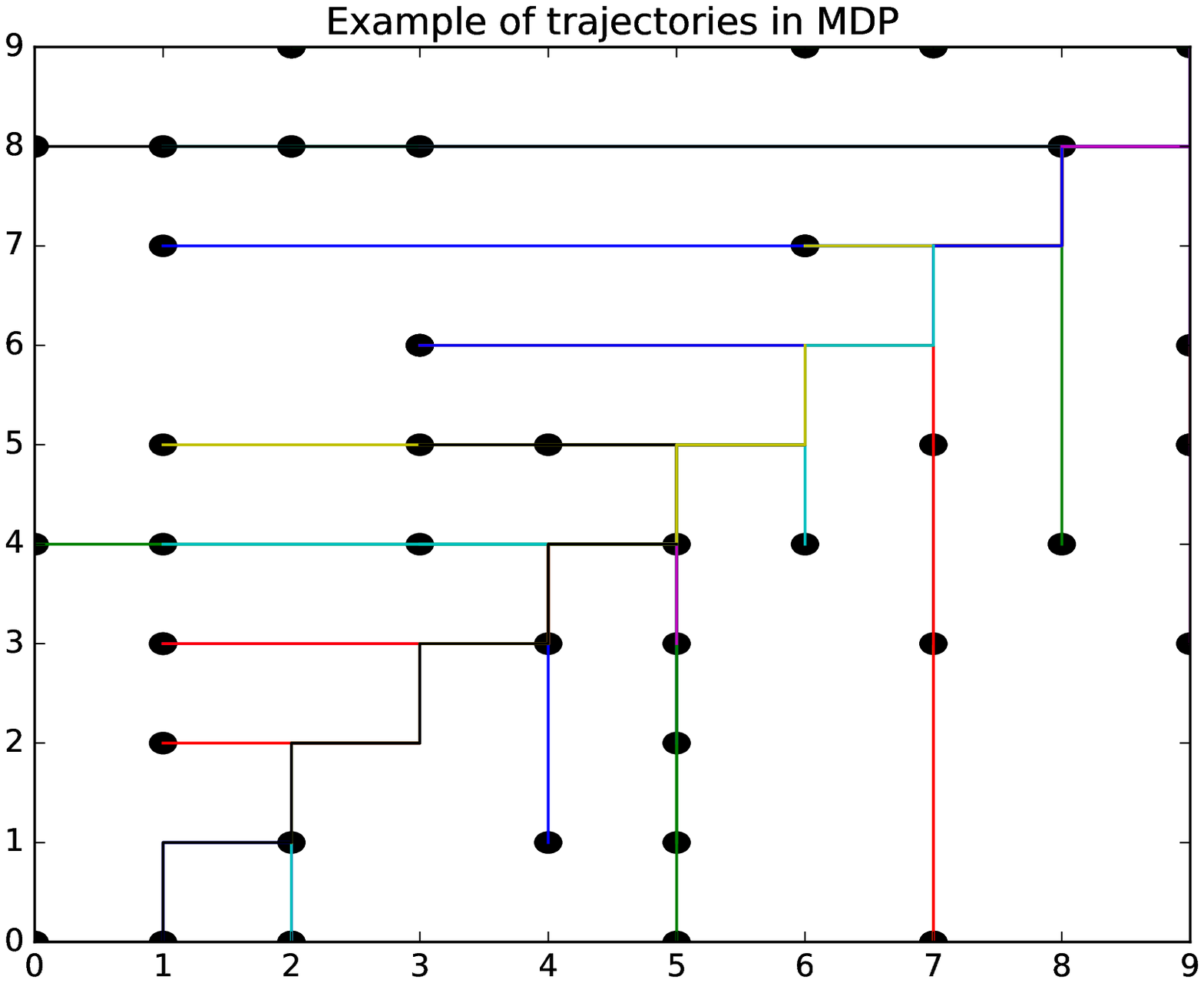}
  \label{fig:environmenttrajectories}}
  \caption{A simulated environment}
\end{figure}

\begin{figure}
  \centering
  \subfloat[A reward table for the gridworld mdp on a $10\times 10$ grid.]{
    \includegraphics[width=0.2\textwidth]{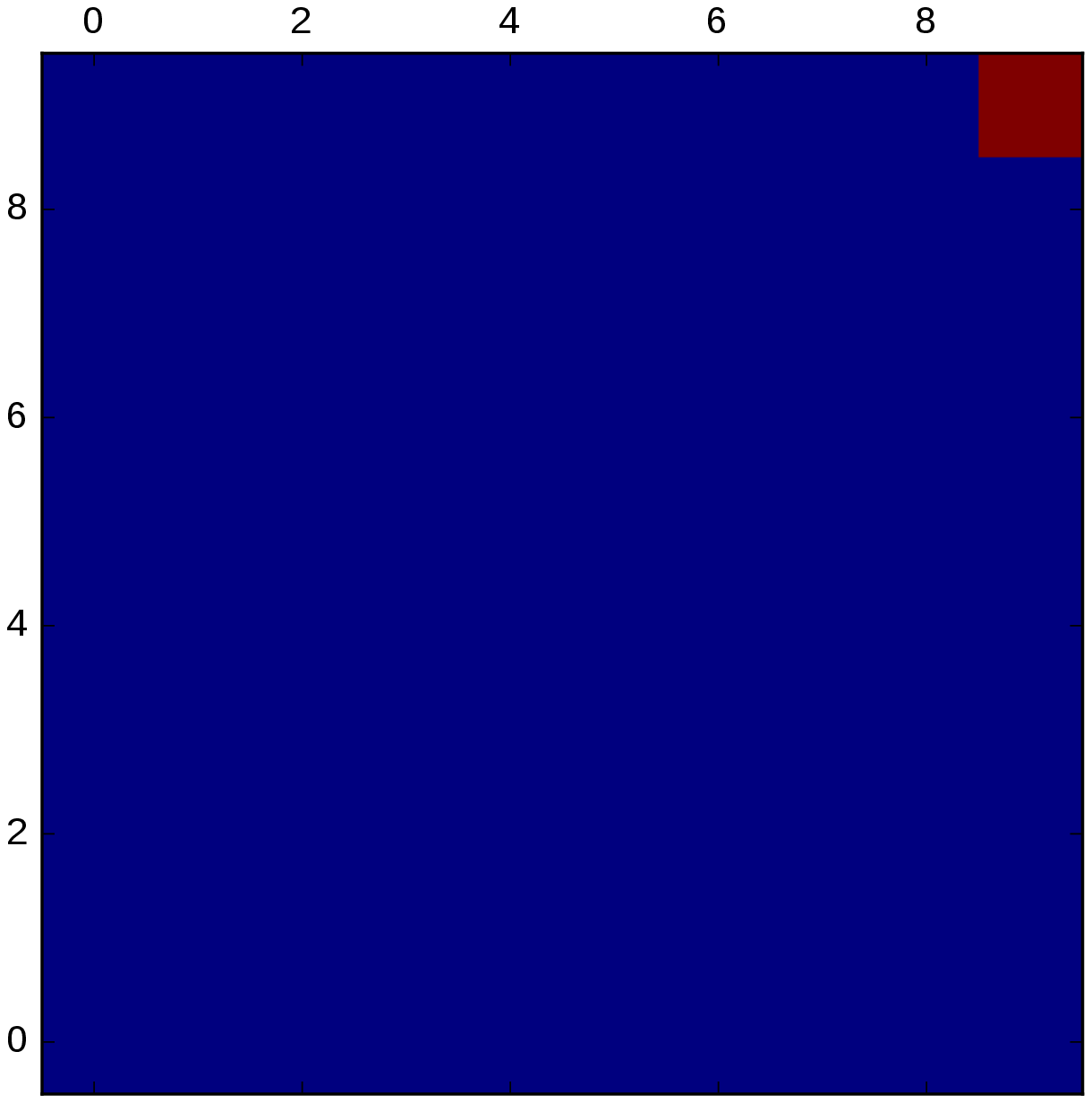}
    \label{fig:gridworld}
  }
  ~
  \subfloat[An example of a reward table for one objectworld mdp on a $10\times 10$ grid: it depends
  on randomly placed objects.]{
    \includegraphics[width=0.2\textwidth]{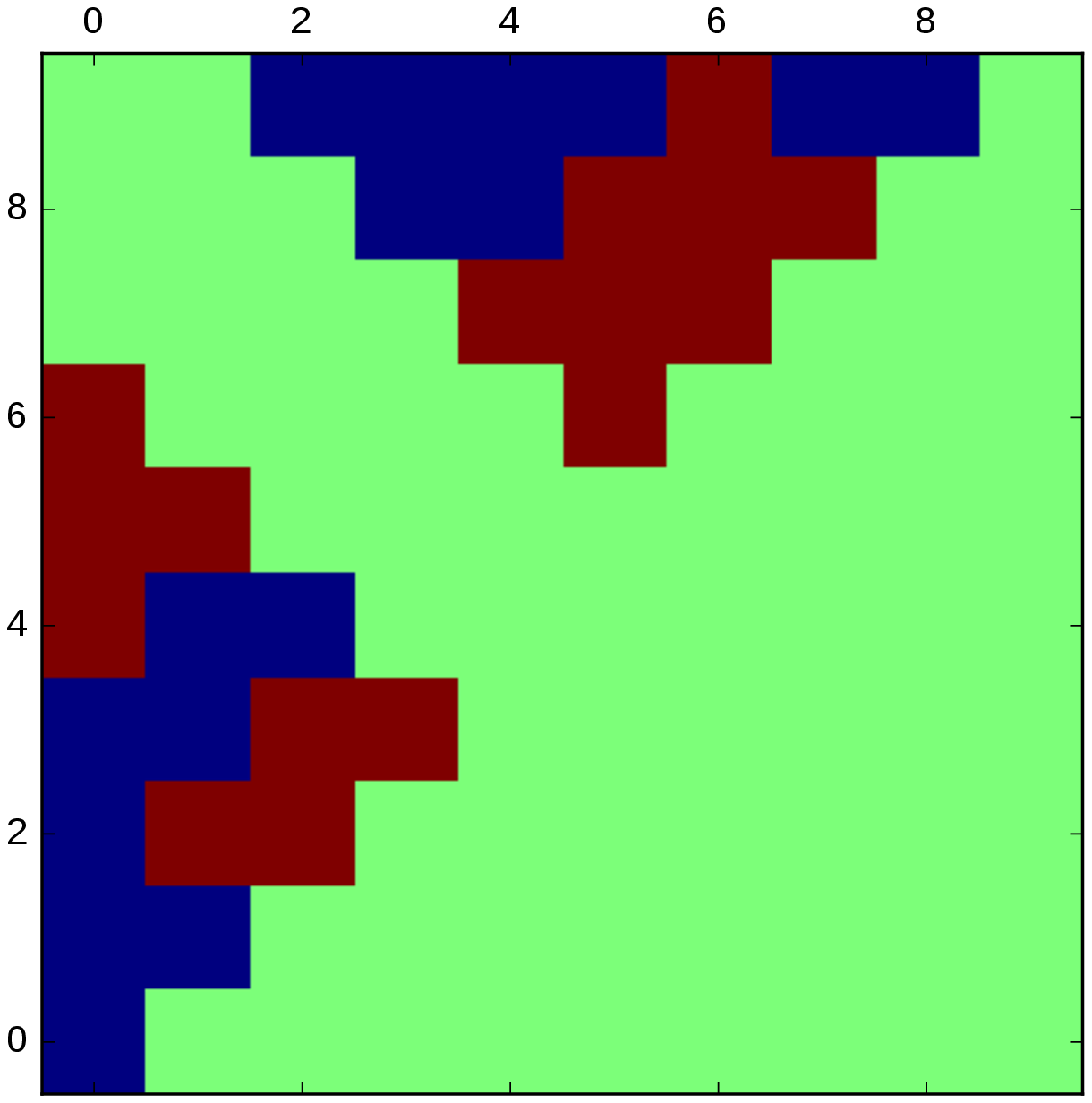}
  \label{fig:objectworld}
}
  \caption{Examples of true reward tables}
\end{figure}

The first example environment is a parking space behind a store, as shown in Figure
\ref{fig:environment}. At each moment, a mobile robot tries to generate an estimation of the
location of the exit based on the observed motions of multiple agents, like cars. Assuming that the
true exit is in one corner of the space, we can describe it with the gridworld mdp \cite{irl::irl1},
and solve the problem via online inverse reinforcement learning. In this $N\times N$ grid, the true
but a-priori unknown rewards for all states equal to zero, except for the upper-right corner state,
whose reward is one, corresponding to the true exit, as shown in Figure \ref{fig:gridworld}.  Each
agent starts from a random state, and chooses in each step one of the following actions: up, down,
left, and right.  Some trajectories are shown in Figure \ref{fig:environmenttrajectories}.  Each
action has a 30\% probability that a random action from the set of actions is actually taken.  For
inverse reinforcement learning, we compare a linear function and a non-linear function to represent
the reward, where the feature of a state is a length-$N^2$ vector indicating the position of the
grid represented by the state, e.g., the $i_{th}$ element of the feature vector for the $i_{th}$
state equals to one and all other elements are zeros. The non-linear function is given as a neural
network with two hidden layers, each with 10 nodes.

The second environment is an objectworld mdp \cite{irl::gaussianirl}. It is similar to the gridworld
mdp, but with a set of objects randomly placed on the grid. Each object has an inner color and an
outer color, selected from a set of possible colors, $C$. The reward of a state is positive if it is
within 3 cells of outer color $C1$ and 2 cells of outer color $C2$, negative if it is within 3 cells
of outer color $C1$, and zero otherwise. Other colors are irrelevant to the ground truth reward. One
example is shown in Figure \ref{fig:objectworld}. This work places two random objects on the grid,
and compares a linear function and a nonlinear function to represent the reward, where the feature
of a state indicates its discrete distance to each inter color and outer color in $C$. The
non-linear function is given as a neural network with two hidden layers, each with 10 nodes.
\subsection{Results}
To evaluate the utility of the proposed algorithm, we use correlation coefficients to measure the
similarity between the learned reward and the true rewards at each moment.

In each environment, 150,000 state-action pairs are generated based on the true rewards. To simulate
a long-term real world observation, we do not assume the environment state to be static if without
robot actions; instead, it is randomly changed every three observations, and the agent has to choose
a sequence of actions reactively.

\begin{figure}
  \centering
  \includegraphics[width=0.4\textwidth]{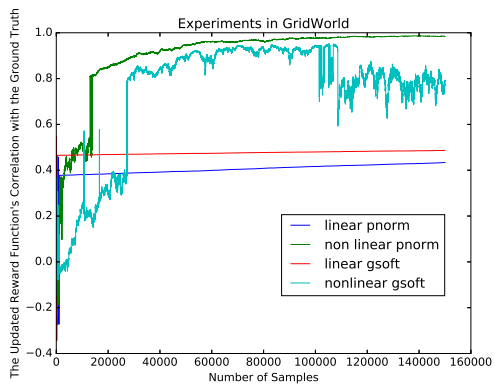}
  \caption{Online inverse reinforcement learning in gridworld: in the figure, "linear" and
  "nonlinear" denote the linear reward function and nonlinear reward function. "pnorm" and "gsoft"
  denote the approximation methods. The correlation coefficient between the learned reward and the
  true reward at each moment is plotted. It shows that the accuracy of the reward function increases
  as the number of samples increases, and the accuracy of nonlinear reward functions increase faster
  than the linear reward functions.}
  \label{fig:onlinegridresult}
\end{figure}

\begin{figure}
  \centering
  \includegraphics[width=0.4\textwidth]{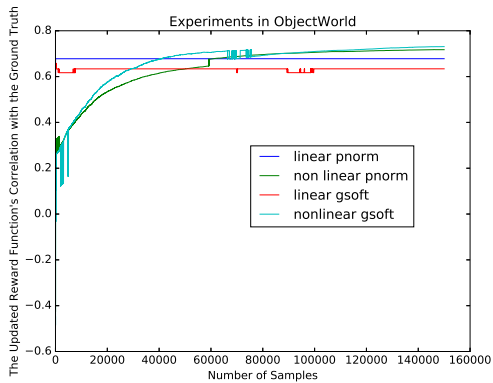}
  \caption{Online inverse reinforcement learning in objectworld: in the figure, "linear" and
  "nonlinear" denote the linear reward function and nonlinear reward function. "pnorm" and "gsoft"
  denote the approximation methods.  The correlation coefficient between the learned reward and the
  true reward at each moment is plotted. It shows that the accuracy of the reward function increases
  as the number of samples increases, and the accuracy of linear reward functions increase faster
  than the nonlinear reward functions.}
  \label{fig:onlineobjectresult}
\end{figure}

We run the proposed algorithm on the data collected from each environment with 30 random
initializations simultaneously, and for each new observation, the correlation coefficient between
the ground truth and the reward function with the highest likelihood among the 30 candidates is
recorded. Besides, we test the algorithm with two approximation methods, pnorm and gsoft, and test
each approximation method with a linear reward function and a nonlinear reward function. For the
linear reward function, we manually choose the learning rate as 0.00001 in both environments, and
for the nonlinear reward function, the learning rate is chosen as 0.001. The approximation
parameters are chosen as $k=100,b=1$. The results are plotted in Figure \ref{fig:onlinegridresult}
and \ref{fig:onlineobjectresult}.

The results show that the accuracy of the learned reward increases as the observation time
increases. While the accuracy reaches the optimum point faster under a linear reward function, the
accuracy is higher under a non-linear reward function.


\subsection{Smart Home Cleaning Robot}
We create a simulated home environment with a person inside and show how the proposed method
improves the efficiency of a home cleaning robot.

Many existing robot cleaners move around the home uniformly, to make sure that every area is evenly
covered. However, this may be inefficient since some areas require more attention while other areas
need less work. To rank different home areas, we assume that more cleaning should be done in areas
which are more frequently visited, and such preference is learned by observing the human activity
via inverse reinforcement learning. Since the preference varies among different users, the cleaning
robot needs to learn it in an online way after being employed.

\begin{figure}
  \centering
  \includegraphics[width=0.4\textwidth]{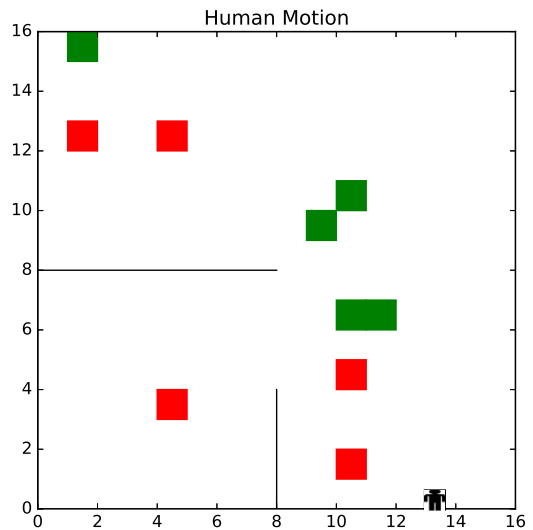}
  \caption{Human movements at home: the home environment is composed of spaces and walls (black
  lines), and the user, represented with the black shape in the figure, moves around at home
  non-uniformly, affected by home objects represented with red and green squares in the figure.}
  \label{fig:smartrobotcleaner}
\end{figure}

The room environment is composed of walls and spaces, and there are furnitures that affect the
preference of the person, as shown in Figure \ref{fig:smartrobotcleaner}. This environment is
discretized into a $16\times16$ grid, where the robot can intermittently observe the movement of the
person, and update its internal reward function based on each new observation. We simulate 5000
human actions, and the robot uses the g-soft approximation with $k=100,b=1$ to learn the reward
function.  A linear reward function and a nonlinear reward function based on a three layer neural
network, with twenty nodes in each layer, are adopted during online inverse reinforcement learning.
Ten initial starting parameters are adopted for each reward function, and at each step, the reward
leading to the highest likelihood is output. The online learned reward function is visualized in the
attached video.

\begin{figure}
  \centering
  \includegraphics[width=0.4\textwidth]{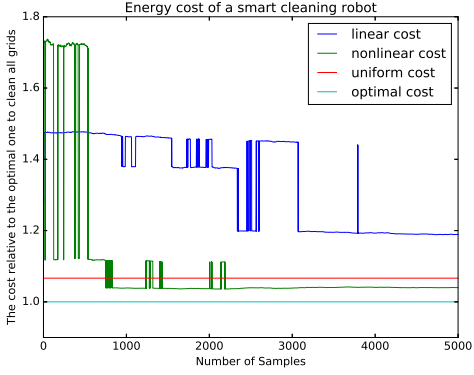}
  \caption{Energy cost of a smart cleaning robot under different strategies: the optimal strategy
  cleans the floor based on the ground truth distribution, and the uniform strategy uses a uniform
  distribution. The proposed method learns a distribution from observed behaviors, and uses it to
  guide the cleaning. In the figure, the cost of the optimal approach is set as 1, and other
  approaches are compared with it.}.
  \label{fig:cleanresult}
\end{figure}

Assuming that the floor's dirt level is distributed as the true reward function and the robot need
to clean all the grids to make the dirt level equal to zero, we evaluate the proposed method based
on the consumed energy in cleaning. The energy cost is computed based on how many times the robot
needs to sweep the whole area, with different dirt distributions, to make every corner clean.  We
compare the accumulated energy consumption by following the uniform cleaning approach, the optimal
approach with the true reward, the proposed approach with a linear reward function, and the proposed
approach with a non-linear reward function. The result in given in Figure \ref{fig:cleanresult}.

It shows that with a non-linear reward function, the cleaning robot consumes less energy than the
typical uniform approach. The amount of saved energy depends negatively on the uniformity of the
true reward function.

\section{Conclusions}
\label{irl::conclusions}
This work formulates an online inverse reinforcement learning algorithm to estimate a reward
functions based on sequentially observed actions of an agent. For each new observation, a predicted
action distribution is computed based on previous reward function, and the reward function is
updated to increase the likelihood of the newly observed action. The action distribution is
formulated as a function of the optimal Q-value, and the gradient of the optimal Q-value with
respect to the reward function is computed via Bellman Gradient Iteration. The algorithm is tested
in two simulated environments based on two approximation methods. The result shows that the proposed
method gradually approaches the true reward function as the number of samples increases, but only
requires limited storage space and computation time. A potential application to home cleaning robots
is demonstrated.

In future work, we will explore different variants of stochastic gradient descent for online inverse
reinforcement learning, and apply the method to several long-term applications, like human motion
analysis.

\appendix
\subsection{P-Norm Approximation}
\label{sec:pnorm}
  Assuming $Q_{p,k}(s,a,\theta_t)\geq 0, \forall s,\forall a$, the tight lower bound of $g_{p,k}(k)$
  is zero: \[\inf_{\forall k\in \mathbb{R}} g_{p,k}(k)=0.\]

\begin{proof}
  $\forall k \in \mathbb{R}$, assuming $a_{max}=\argmax_{a'} Q_{p,k}(s',a',\theta_t)$,
    \begin{align}
      g_{p,k}(k)&=(\sum_{a'\in A} Q_{p,k}(s',a',\theta_t)^k)^{\frac{1}{k}}-\max_{a'\in A}Q_{p,k}(s',a',\theta_t)\nonumber\\
      &=(\sum_{a'\in
      A/a_{max}}Q_{p,k}(s',a',\theta_t)^k+Q_{p,k}(s',a_{max},\theta_t)^k)^{\frac{1}{k}}\nonumber\\
      &\quad-\max_{a'\in A}Q_{p,k}(s',a',\theta_t)\nonumber.
    \end{align}
  Since $Q_{p,k}(s,a,\theta_t)\geq 0\Rightarrow\sum_{a'\in A/a_{max}}Q_{p,k}(s',a',\theta_t)^k\geq 0$,
    \begin{align}
      g_{p,k}(k)&\geq(Q_{p,k}(s',a_{max})^k,\theta_t)^{\frac{1}{k}}-\max_{a'\in
      A}Q_{p,k}(s',a',\theta_t)\nonumber\\
      &=Q_{p,k}(s',a_{max},\theta_t)-\max_{a'\in A}Q_{p,k}(s',a',\theta_t)=0\nonumber
    \end{align}
   When $k=\infty$:
    \begin{align}
      g_{p,k}(k)&=(\sum_{a'\in A} Q_{p,k}(s',a',\theta_t)^\infty)^{\frac{1}{\infty}} -\max_{a'\in A}Q_{p,k}(s',a',\theta_t)\nonumber\\
      &=\max_{a'\in A}Q_{p,k}(s',a',\theta_t)-\max_{a'\in A}Q_{p,k}(s',a',\theta_t)=0\nonumber
    \end{align}
\end{proof}
Assuming all Q-values are non-negative, $Q_{p,k}(s,a,\theta_t)\geq 0, \forall s,a$, $g_{p,k}(k)$ is
a decreasing function with respect to increasing $k$: \[g'_{p,k}(k)\leq0, \forall k \in
\mathbb{R}.\]

\begin{proof}
  \begin{align}
    g'_{p,k}(k)&=\frac{1}{k}*(\sum_{a'\in A}
      Q_{p,k}(s',a',\theta_t)^k)^{\frac{1-k}{k}}*(\nonumber\\
      &\sum_{a'\in A}
      Q_{p,k}(s',a',\theta_t)^k\log(Q_{p,k}(s',a',\theta_t)))\nonumber\\
      &+(\sum_{a'\in A} Q_{p,k}(s',a',\theta_t)^k)^{\frac{1}{k}}\log(\sum_{a'\in A}
      Q_{p,k}(s',a',\theta_t)^k)\frac{1}{-k^2}\nonumber\\
      &=\frac{(\sum_{a'\in A} Q_{p,k}(s',a',\theta_t)^k)^{\frac{1}{k}}}{ k^2\sum_{a'\in
      A}Q_{p,k}(s',a',\theta_t)^k}(\nonumber\\
      & \sum_{a'\in
      A}Q_{p,k}(s',a',\theta_t)^kk\log(Q_{p,k}(s',a',\theta_t))\nonumber\\
      &- \sum_{a'\in A}Q_{p,k}(s',a',\theta_t)^k \log(\sum_{a'\in A}Q_{p,k}(s',a',\theta_t)^k))\nonumber.
  \end{align}
  Since $k\log(Q_{p,k}(s',a',\theta_t))\leq\log(\sum_{a'\in A}Q_{p,k}(s',a',\theta_t)^k)$:
  \[g_{p,k}'(k)\leq0.\]
\end{proof}

\subsection{G-Soft Approximation}
\label{sec:gsoft}
  The tight lower bound of $g_{g,k}(k)$ is zero:
  \[\inf_{\forall k\in \mathbb{R}} g_{g,k}(k)=0.\]

\begin{proof}
  $\forall k\in \mathbb{R}$: assuming $a_{max}=\argmax_{a'}Q_{g,k}(s',a',\theta_t), $
    \begin{align}
      &g_{g,k}(k)=\frac{\log(\sum_{a'\in A}\exp(kQ_{g,k}(s',a',\theta_t)))}{k}-\max_{a'\in A}Q_{g,k}(s',a',\theta_t)\nonumber\\
      &=\frac{\log(\sum_{a'\in A/a_{max}}\exp(kQ_{g,k}(s',a',\theta_t))+\exp(kQ_{g,k}(s',a_{max},\theta_t)))}{k}\nonumber\\&
      -\max_{a'\in A}Q_{g,k}(s',a',\theta_t)\nonumber\\
      &>Q_{g,k}(s',a_{max},\theta_t)-\max_{a'\in A}Q_{g,k}(s',a',\theta_t)=0\nonumber
    \end{align}
    When $k=\infty$,
    \begin{align}
      &\lim_{k\to\infty}(\frac{\log(\sum_{a'\in A}\exp(kQ_{g,k}(s',a',\theta_t)))}{k}-\max_{a'\in
      A}Q_{g,k}(s',a',\theta_t))\nonumber\\
      &=\lim_{k\to\infty}(\frac{\log(\sum_{a'\in A}\exp(kQ_{g,k}(s',a',\theta_t)))}{k})-\max_{a'\in
      A}Q_{g,k}(s',a',\theta_t)\nonumber\\
      &=\max_{a'\in A}Q_{g,k}(s',a',\theta_t)-\max_{a'\in A}Q_{g,k}(s',a',\theta_t)=0\nonumber
    \end{align}
\end{proof}

 $g_{g,k}(k)$ is a decreasing function with respect to increasing $k$:
  \[g'_{g,k}(k)<0, \forall k \in \mathbb{R}.\]

\begin{proof}
  \begin{align}
    g'_{g,k}(k)&=-\frac{\log(\sum_{a'\in A}\exp(kQ_{g,k}(s',a',\theta_t)))}{k^2}\nonumber\\
      &+\frac{\sum_{a'\in A}Q_{g,k}(s',a',\theta_t)\exp(kQ_{g,k}(s',a',\theta_t))}{k\sum_{a'\in
      A}\exp(kQ_{g,k}(s',a',\theta_t))}<0\nonumber
  \end{align}
    Since:
    \begin{align}
      &-\frac{\log(\sum_{a'\in A}\exp(kQ_{g,k}(s',a',\theta_t)))}{k^2}\nonumber\\
      &+\frac{\sum_{a'\in A}Q_{g,k}(s',a',\theta_t)\exp(kQ_{g,k}(s',a',\theta_t))}{k\sum_{a'\in A}\exp(kQ_{g,k}(s',a',\theta_t))}<0\nonumber\\
      &\Longleftrightarrow  \sum_{a'\in A}kQ_{g,k}(s',a',\theta_t)\exp(kQ_{g,k}(s',a',\theta_t))<\nonumber\\
      &\sum_{a'\in A}\log(\sum_{a'\in A}\exp(kQ_{g,k}(s',a',\theta_t)))\exp(kQ_{g,k}(s',a',\theta_t))\nonumber\\
      &\Longleftarrow kQ_{g,k}(s',a',\theta_t)<\log(\sum_{a'\in A}\exp(kQ_{g,k}(s',a',\theta_t)))\nonumber.
    \end{align}
\end{proof}

\subsection{Convergence Analysis}
\label{sec:convergence}
  Assuming $\theta^*$ is a local optimum of the objective function under the true Q function and
  $L(\theta^*,s,a)$ is the locally optimal value, the gradient method
  $\theta_{t+1}=\theta_t+\alpha*\nabla L(\theta_t,k,s,a)$ from a starting point in basin of a
  $\theta^*$ under the approximated gradient $\nabla L(\theta_t,k,s,a)$ will converge to
  $\theta^{k,*}$, $\forall \epsilon<<1,\exists k$ such that
  $||L(\theta^*,s,a)-L(\theta^{k,*},s,a)||<\epsilon$ and $\lim_{k\rightarrow\infty}\epsilon=0$.

\begin{proof}
  First, we show that the gradient method will converge to a point under the approximated gradient.
  We consider the case when the objective function is defined on one state-action pair, but the
  result can be easily applied to general cases.

  Stochastic gradient methods update the reward parameter once for each observation. For a update
  step $\alpha*\nabla L_k(\theta_t,s,a)$, we expand the updated objective function with first-order
  Taylor series:
  \begin{align}
    L(\theta_{t+1},s,a)&=L(\theta_t+\alpha*\nabla L_k(\theta_t,s,a))\nonumber\\
                   &\approx L(\theta_t,s,a)+\alpha*\nabla L(\theta_t,s,a)\cdot\nabla
                   L_k(\theta_t,s,a)
    \label{equation:taylor}
  \end{align}
  where
  \begin{align}
    \nabla L(\theta_t,s,a)&=b*\nabla Q(s,a,\theta_t)\nonumber\\
    &-b*\sum_{\tilde{a}\in
    A}P(\tilde{a}|\theta_t;s)\nabla Q(s,\tilde{a},\theta_t),
    \label{equation:maxgradient}
  \end{align}
  \begin{align}
    \nabla L_k(\theta_t,s,a)&=b*\nabla Q_k(s,a,\theta_t)\nonumber\\
    &-b*\sum_{\tilde{a}\in
    A}P(\tilde{a}|\theta_t;s)\nabla Q_k(s,\tilde{a},\theta_t).
    \label{equation:approxgradient}
  \end{align}
  $\nabla Q_k(s,a,\theta_t)$ denotes the approximate gradient under \textit{p-norm
  approximation} or \textit{g-soft approximation}.

  Due to the max-operation in Bellman Optimality Equation, $\nabla Q(s,a,\theta_t)$ is a piecewise
  smooth function of $\theta_t$, and the gradient is defined on the optimal paths following $(s,a)$
  under current $\theta_t$, while $\nabla Q_k(s,a,\theta_t)$ is defined on both the optimal path and
  all the other non-optimal paths following $(s,a)$. We describe $\nabla Q_k(s,a,\theta_t)$ as a
  weighted summation of the optimal paths and the non-optimal paths, where the weight is a function
  of $k$:
  \begin{align}
    \nabla Q_k(s,a,\theta_t)= &w_A(k,\theta_t)*\nabla Q(s,a,\theta_t)+\nonumber\\
     &w_B(k,\theta_t))*\nabla Q(s_r,a_r,\theta_t)
    \label{equation:weightgrad}
  \end{align}
  where $(s_r,a_r)$ denotes the state action
  pairs in the non-optimal paths, and
  \[w_A(k,\theta_t)=\frac{\partial Q_k(s,a,\theta_t)}{\partial Q(s,a,\theta_t)},\]
  \[w_B(k,\theta_t)=\frac{\partial Q_k(s,a,\theta_t)}{\partial Q(s_r,a_r,\theta_t)}.\]
  As $k$ increases, $Q_k(s,a,\theta_t)$ will depends more on the optimal paths, thus $w_A(k,\theta_t)$
  will increase while $w_B(k,\theta_t)$ will decrease. When $k\rightarrow\infty$,
  $w_A(k,\theta_t)\rightarrow 1$ and $w_B(k,\theta_t)\rightarrow 0$.

  This description is reasonable based on Equation \eqref{equation:pnormvgrad},
  \eqref{equation:pnormqgrad} \eqref{equation:gsoftvgrad}, and \eqref{equation:gsoftqgrad}, where in
  each iteration, the approximate gradient of a state-action pair is defined as a weighted summation
  of the gradients of the resultant state-action pairs, and the weights of the optimal ones approach
  1 as $k$ increases, thus the final approximated gradient can be described as a weighted summation
  of the optimal sequences of state-action pairs and other sequences.

  Substituting Equation \eqref{equation:weightgrad} into Equation \eqref{equation:taylor}:
  \begin{align}
    L(\theta_{t+1},s,a)&=L(\theta_t+\alpha*\nabla L_k(\theta_t,s,a))\nonumber\\
                   &\approx L(\theta_t,s,a)+\alpha*(w_A(k,\theta_t)*\nabla L(\theta_t,s,a)\cdot\nabla
                   L(\theta_t,s,a)\nonumber\\
                   &\qquad+w_B(k,\theta_t)*\nabla L(\theta_t,s,a)\cdot\nabla L(\theta_t,s_r,a_r)).
    \label{equation:taylor1}
  \end{align}

Assuming $||\nabla L(\theta_t,s,a)||=N_L,||\nabla L(\theta_t,s_r,a_r)||=N_r$, in the best case, if
the two gradients have the same direction,  the improvement is
$\alpha*(w_A(k,\theta_t)*N_L^2+w_B(k,\theta_t)*N_r*N_L)$; in the worst case, if they have the
opposite directions,  the improvement is $\alpha*(w_A(k,\theta_t)*N_L^2-w_B(k,\theta_t)*N_r*N_L)$.
Therefore, the improvement of the objective function depends on
$\frac{w_A(k,\theta_t)}{w_B(k,\theta_t)}$. If
$\frac{w_A(k,\theta_t)}{w_B(k,\theta_t)}>\frac{N_r}{N_L}$ for a sufficiently large $k$, we can make
sure that the objective function is converging:
  \[L(\theta_{t+1},s,a)>L(\theta_t,s,a).\]
  The convergence rate depends on $w_A(k,\theta_t)$, $w_B(k,\theta_t)$, and the MDP structure.

Second, we show that $\forall \epsilon,\exists k>N \text{ such that }
||L(\theta^*)-L(\theta^{k,*})||<\epsilon$.

We construct an approximate objective function $L_k(\theta_t,s,a)$ by replacing the Q value in
Equation \eqref{equation:loglikelihood} with an approximated Q value:
\[L_k(\theta_t,s,a)=b*Q_k(s,a,\theta_t)-\log{\sum_{\tilde{a}\in
  A}\exp{b*Q_k(s,\tilde{a},\theta_t)}}
\]
where $Q_k(s,a,\theta)$ denotes the approximate gradient under \textit{p-norm
  approximation} or \textit{g-soft approximation}. $L_k(\theta_t,s,a)$ is a
differentiable equation and a local optimum $\theta^{k,*}$ can be reached via gradient method.

Based on proposition 1, 2, 3, and 4, we propose that $\forall \eta, \forall \theta,
\exists k \text{ such that } ||Q(s,a,\theta)-Q_k(s,a,\theta)||<\eta$. Thus the following inequality
holds:
\begin{align}
  ||L(\theta_t,s,a)-L_k(\theta_t,s,a)||&=||Q(s,a,\theta_t)-Q_k(s,a,\theta_t)-\nonumber\\
  &\log{\frac{\sum_{\tilde{a}\in A}\exp{b*Q(s,\tilde{a},\theta_t)}}{\sum_{\tilde{a}\in A}\exp{b*Q_k(s,\tilde{a},\theta_t)}}}|| \nonumber\\
  &\leq ||Q(s,a,\theta_t)-Q_k(s,a,\theta_t)||+\nonumber\\
  &||\log{\frac{\sum_{\tilde{a}\in
  A}\exp{b*Q(s,\tilde{a},\theta_t)}}{\sum_{\tilde{a}\in
  A}\exp{b*Q_k(s,\tilde{a},\theta_t)}}}|| \nonumber\\
  &\leq \eta+b*\eta=(1+b)*\eta.
\end{align}

Therefore, $\forall \epsilon$, we may choose a $k>N$ leading to $\eta\leq \frac{\epsilon}{1+b}$.
This $k$ will guarantee $||L(\theta_t,s,a)-L_k(\theta_t,s,a)||<\epsilon$ for any $\theta$. Since
$\theta^*$ and $\theta^{k,*}$ represent two close local optimum points, we deduce that
$||L(\theta^*,s,a)-L(\theta^{k,*},s,a)||\leq||L(\theta^*,s,a)-L_k(\theta^*,s,a) ||\leq\epsilon$.

Wih Taylor expansion,
\[L(\theta^*,s,a)=L(\theta^{k,*},s,a)+L'(\theta^{k,*},s,a)(\theta^*-\theta^{k,*}).\]
Thus,
\[||L'(\theta^{k,*},s,a)||||(\theta^*-\theta^{k,*})||=||L(\theta^*,s,a)-L(\theta^{k,*},s,a)||\leq
\epsilon,\]
and,
\[||(\theta^*-\theta^{k,*})||\leq\frac{\epsilon}{||L'(\theta^{k,*},s,a)||}.\]
\end{proof}

\end{document}